\documentclass[letterpaper, 10 pt, conference]{ieeeconf}
\IEEEoverridecommandlockouts
\usepackage{float}
\usepackage{graphicx}
\usepackage{amssymb}

\usepackage{amsmath}
\usepackage{mathtools}  

\usepackage{nccmath}

\usepackage{amsthm}
\usepackage[usenames,dvipsnames,svgnames,table]{xcolor}
\theoremstyle{plain}

\newtheorem{proposition}{Proposition}
\newtheorem{definition}{Definition}

\newtheorem{problem}{Problem}

\usepackage{float}
\floatstyle{plaintop}
\restylefloat{table}
\usepackage{MnSymbol}%
\usepackage{wasysym}%
\usepackage{caption}
\usepackage{subcaption}
\usepackage{graphicx}
\usepackage{import}

\usepackage[hidelinks]{hyperref}

\usepackage{algorithm}
\usepackage{algpseudocode}

\renewcommand{\baselinestretch}{0.945}

\definecolor{deepblue}{rgb}{0,0,1}

\begin{document}

\title{ 
\bf 
Folding Knots Using a Team of Aerial Robots
}

\author{Diego S. D'Antonio and David Salda\~{n}a
\thanks{$^1$The authors are with the Autonomous and Intelligent Robotics Laboratory --AIRLab-- at Lehigh University, Bethlehem, PA, 18015, USA. \texttt{Email:~\{diego.s.dantonio,~saldana\}@lehigh.edu}}
\thanks{$^2$The authors would like to thank Gustavo A. Cardona, Jiawei Xu, and Disha Kamale at Lehigh University for their useful and constructive comments on this project.
}
}


\maketitle

\begin{abstract}

From ancient times, humans have been using cables and ropes to tie, carry, and manipulate objects by folding knots. However, automating knot folding is challenging because it requires dexterity to move a cable over and under itself. In this paper, we propose a method to fold knots in midair using a team of aerial vehicles. 
We take advantage of the fact that vehicles are able to fly in between cable segments without any re-grasping. So the team grasps the cable from the floor, and releases it once the knot is folded.
Based on a composition of catenary curves, we simplify the complexity of dealing with an infinite-dimensional configuration space of the cable, and formally propose a new knot representation.
Such representation allows us to design a trajectory that can be used to fold knots using a leader-follower approach. We show that our method works for different types of knots in simulations. Additionally, we show that our solution is also computationally efficient and can be executed in real-time.

\end{abstract}


\IEEEpeerreviewmaketitle



\section{introduction} 




More recently, aerial robotics have become a topic of great interest in industry and academia because they offer low-cost solutions to a large number of applications, especially in solving problems that involve object manipulation. Aerial vehicles such as quadrotors can transport medicines, deliver food~\cite{bamburry2015drones}, conduct search-and-rescue missions~\cite{cacace2016control}, and explore and effectively map environments . Additionally, improvements in onboard computing power and battery efficiency have made quadrotors advantageous to performing manipulation tasks.


 Aerial robots can physically interact with objects using different mechanisms such as grippers, arms, and cables~\cite{Ollero2022}. A cable, or rope, is a versatile and low-cost tool that can deform its shape depending on the interaction.
 The cable is lightweight, making it an excellent option for aerial applications. However, the dimension of its configuration space is infinite, making it challenging to simulate and control.
 Recent works discretize the cable and analyze its dynamics as a chain \cite{Kotaru2020}, but still computationally slow for real-time applications.
A cable can also be used to build tensile structures, such as bridges using multiple quadrotors \cite{Augugliaro2013}.
 By hanging a cable by its ends, the gravity force pulls it down, forming a catenary curve. 
 The catenary curve has been studied since the sixteen century, and its mathematical framework is well defined~\cite{Laranjeira2017}. Analyzing and controlling the cable configuration in this curve reduces its complexity to a five-dimensional space \cite{catenaryrobot}. Later it has been applied to multiple applications in robotics that ranges from obstacle avoidance with a hanging cable \cite{sreenath2013trajectory}, and visual servoing approaches~\cite{Galea2017, Xiao2018, Abiko2017}.
 In our previous work, we studied the catenary curve, and we found several applications to control the catenary curve through its hanging points~\cite{catenaryrobot}, manipulate cuboid objects~\cite{GustavoCatenary}, and lift objects using adaptive control to overcome uncertainty in contact points and unknown physical parameters \cite{GustavoCatenaryAdaptive}.

From ancient times, humans have been using cables and ropes to carry, tie, and manipulate physical objects by folding knots. However, automating knot folding is even more challenging than cable manipulation because it requires dexterity to move the cable over and under other cable segments.
Knots have been extensively studied as a branch of topology, proving a solid theory that can be used in many applications. 
The necessity to automate the process in medicine, surgical operations and car industry have pushed the application of knot theory in robotics.
{Despite folding knots has been studied using arm robots, those approaches require to} grasp and re-grasp the cable multiple times during the folding process \cite{Wakamatsu2004, Wang2018, Matthew2014}. Other approaches are focused on  topological planning~\cite{bhattacharya2018, saha}. and learning \cite{yan2020learning, Vinh2012}. However, folding knots using autonomous robots is still an active research field because it involves complex dynamics and high dexterity.

\begin{figure}[t]
    \centering
    \includegraphics[trim=.4cm 0cm 0cm 0cm,clip, width=1\linewidth, ]{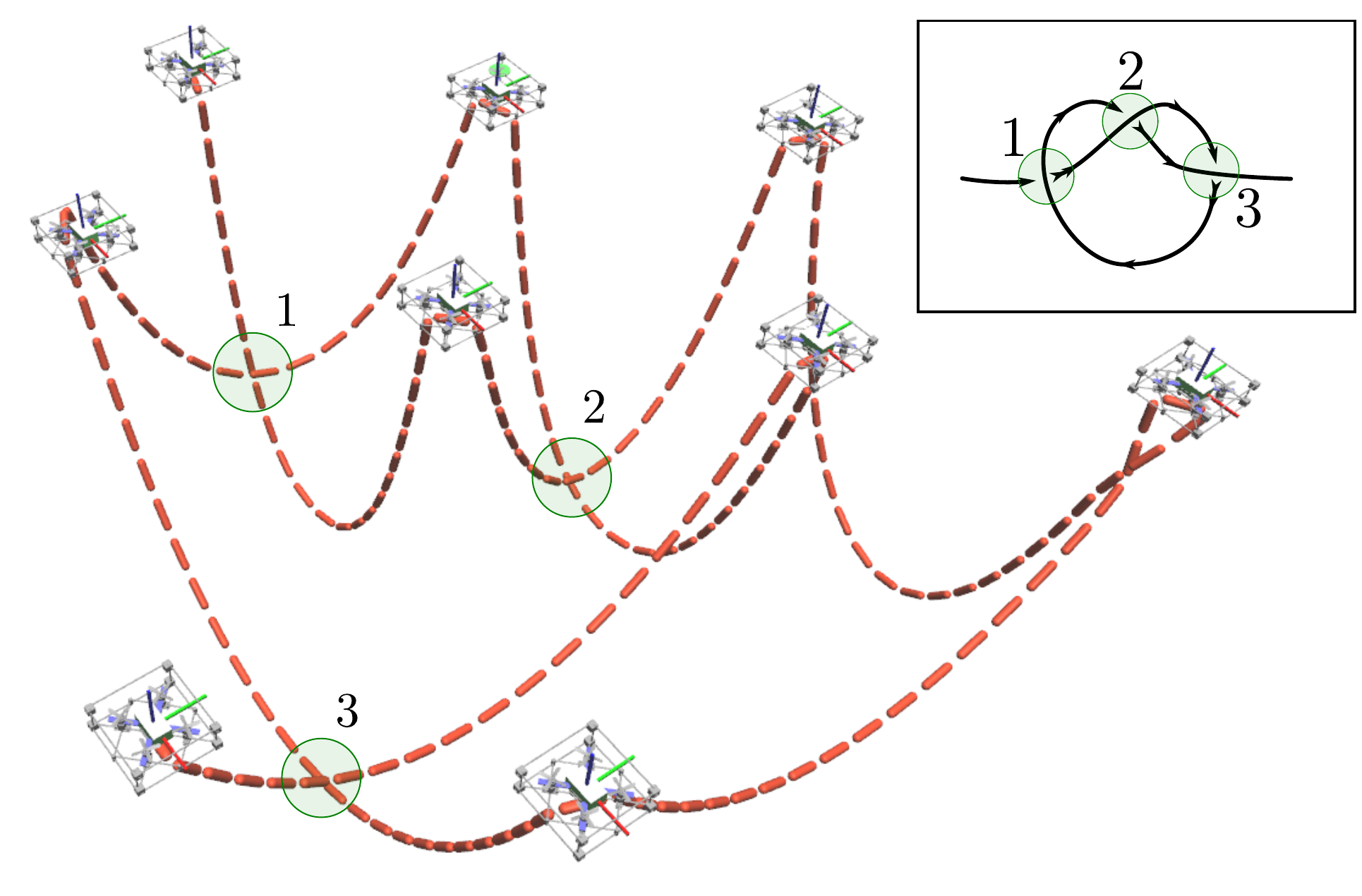}
    \caption{A team of aerial robots folding a knot. The green disks highlight that the three crossings of the desired knot (top right corner) are the same as the actual knot.}
    \label{fig:intro}
\end{figure}





In this work, we propose a method to fold knots in midair using multiple aerial robots. One of the main advantages of using aerial robots is that robots can move between cable segments without requiring re-grasping. Initially, robots grasp the cable from the floor and release it once the knot is folded. We extend an existent knot representation to develop a new representation based on catenary curves. The use of catenary curves simplifies the complexity of dealing with an infinite-dimensional configuration space. This new representation guides the robots to grab the cable and fold it using a leader-follower approach. We illustrate a team of nine quadrotors folding a knot in Fig.~\ref{fig:intro}.




The main contribution of this paper is threefold. First, we propose the multi-catenary robot; a robot that controls multiple catenaries to reduce the manipulation complexity of a cable. Second, we develop a new knot representation based on the combination of multiple catenaries and formally show that our new representation maintains the knot topology. This representation is designed for aerial robots, but it can also be used in other scenarios, e.g.,  underwater. Third, based on the new representation, we propose a leader-follower approach to fold knots.

\section{Problem Statement}
\label{section: Problem Statement}

\begin{figure}[b]
    \centering
    \def\svgwidth{0.45\textwidth}
    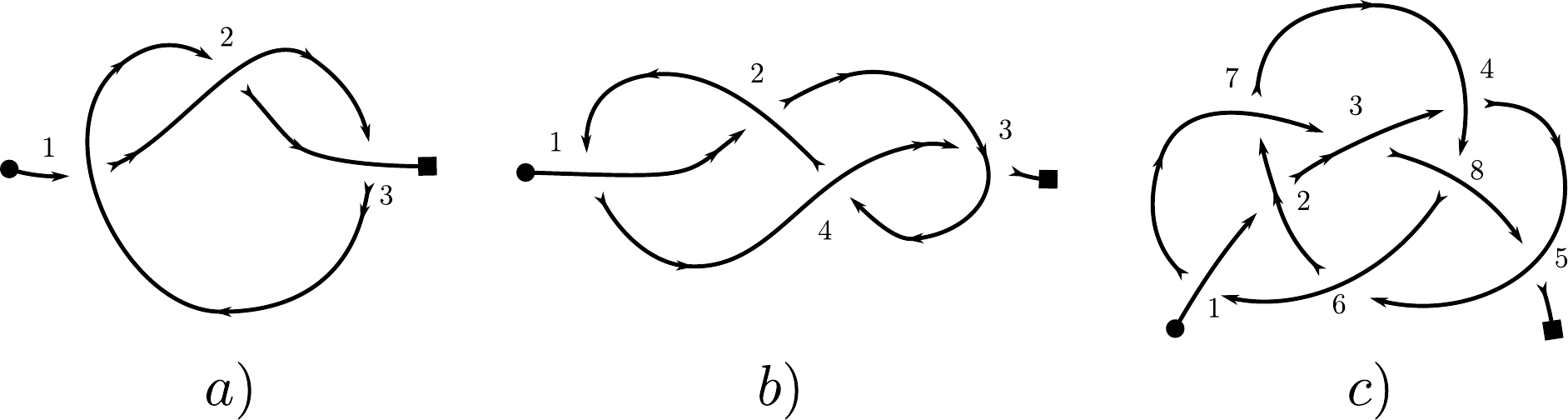
    \caption{Knot examples: a) Overhand knot $c(K)=3$, b) Figure-eight knot $c(K)=4$, c) Carrick mat knot with $c(K)=8$. }
    \label{fig:examples_knot}
\end{figure}

We propose a new way to fold knots in mid-air using a team of aerial robots attached to a cable. We leverage the fundamentals of knot theory~\cite{adams1994knot} to define and represent the target knot.
\begin{definition}[\textbf{Knot}]
A \emph{knot} $K$, is defined as a closed, non-self-intersecting curve that is embedded in the three-dimensional Euclidean space $\mathbb{R}^3$ and cannot be untangled to produce a simple loop (i.e., a simple closed curve). 
\end{definition}
Fig.~\ref{fig:examples_knot} illustrates some examples of knots that we want to fold.
To describe the knot, we use the number of crossing points, $c(K)$, which is the least number of crossings that occur in any projection of the knot to identify the number of crossing points. We start counting and enumerating each self-crossing.
A knot preserves its topology if it is deformed without cutting or pasting. Thus, we can continuously deform a knot and still maintain its topology.

\begin{figure}[t]
    \centering
    \def\svgwidth{0.9\linewidth}
    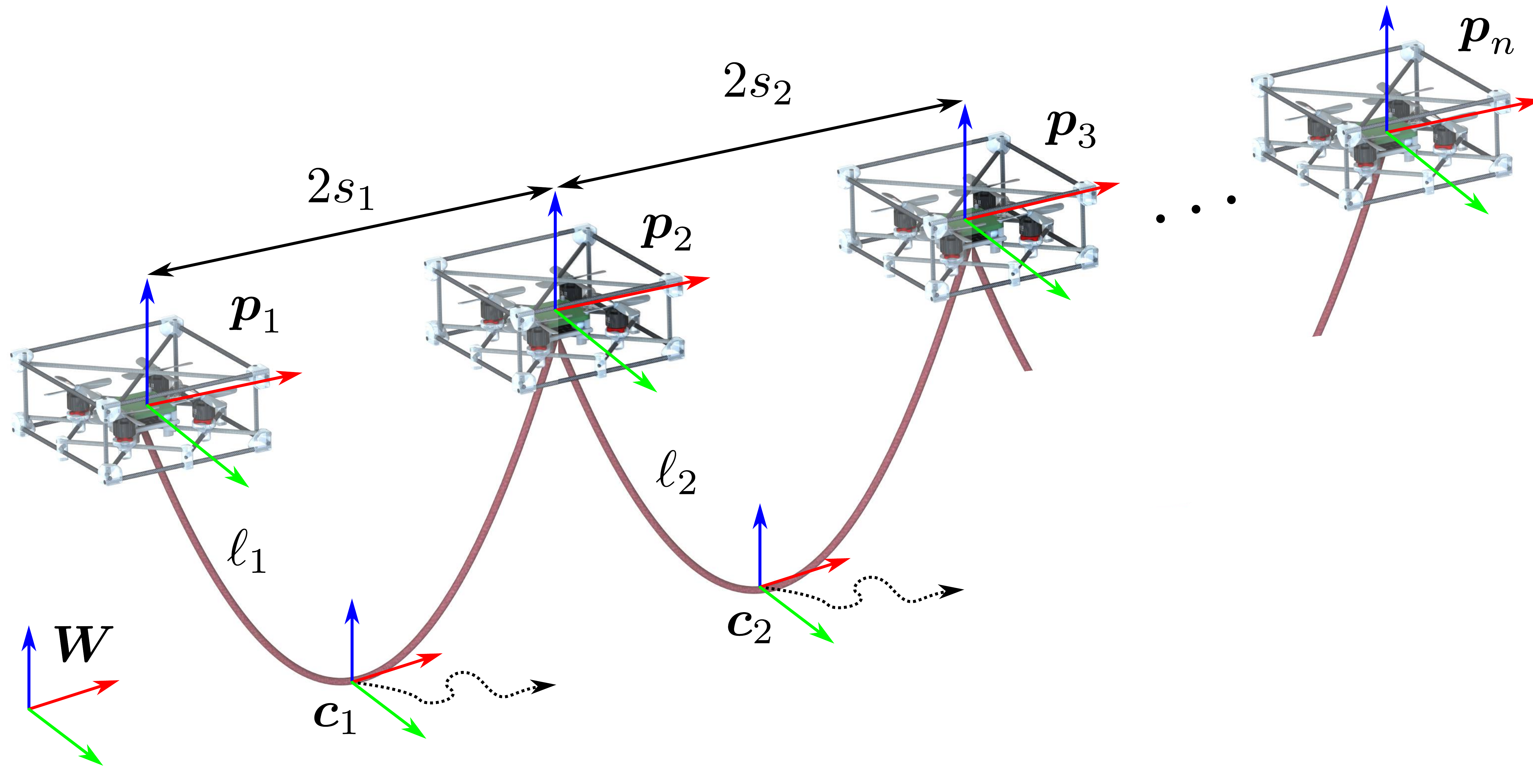
    \caption{A Multi-catenary robot formed by $n$ aerial robots attached to a cable.}
    \label{fig:multi_catenary}
\end{figure}

To fold a knot $K$, we use a team of $n$ robots, where $n$ varies depending on the knot type.
The robots can attach to a cable, forming a line configuration. We label the robots in sequencing order.
The cable has a total length of $\ell_T$.
The location of the $i$th robot is denoted by $\boldsymbol{p}_{i} \in \mathbb{R}^3$, subject to $\|\boldsymbol{p}_{i}-\boldsymbol{p}_{i+1}\|\leq \ell_i$, where $\ell_i$ is the cable length of the segment between robot $i$ and robot $i+1$ (see Fig.~\ref{fig:multi_catenary}). The total length can be expressed as the sum of the cable segments, i.e., $\ell_T~=~\sum_{i=1}^{n-1}\ell_i$.
Part of this work is finding a sufficient number of robots, cable's total length, and cable segments to allow the team to fold the knot.

The robot's motion follows a second-order dynamics, and we assume that we can control their acceleration, $\boldsymbol{\ddot{p}}_i=\boldsymbol{u}$, where $\boldsymbol{u}$ is the control input.
We provide the dynamics model of holding a cable using two quadrotors in our previous work~\cite{catenaryrobot}. This paper focuses on knot folding and assumes that the robots are omnidirectional.

Every pair of robots $(i, i+1)$, for $i=\{1,..., n-1\}$ holds a portion of the cable from two points. This portion of the cable forms a Catenary curve
that starts at point $\boldsymbol{p}_i$ and ends at point $\boldsymbol{p}_{i+1}$.
The array of connected Catenary curves and the robots holding it forms a compound robot called the \emph{Multi-Catenary robot} (see Fig~\ref{fig:multi_catenary}).
\begin{definition}[\textbf{Multi-Catenary Robot}]
\label{def:flying-rope}
A \emph{Multi-Catenary robot} is composed of a set of $n$ aerial robots attached to a non-stretchable cable of length $\ell_T$.
\end{definition}
A multi-catenary robot is a convenient way to manipulate a cable since we can divide the cable in $n-1$ portions and move it as a single multi-link system by controlling the position of the joints.
However, there are two fundamental problems to solve. We need to represent the knot in a way that a multi-catenary robot can form and a method to fold the knot.
\begin{problem}
Design a knot representation based on a multi-catenary curve such that it is homotopic to any other knot representation. The curve should not have any self-intersection.
\end{problem}
\begin{problem}
Given a knot $K$ in any representation, design a method to fold the knot 
using a {multi-catenary robot}.
\end{problem}

\section{Multi-Catenary Representation for Knots} 
\label{section: Knot representation}


In knot theory, there are multiple ways to represent a knot.
One can describe the knot by using the number of crossing $c(K)$ with a tabular representation, e.g., Dowker–Thistlethwaite (DT) notation, Gauss code, Arc representation. For instance, the Gauss code is a knot representation that labels and enumerate the crossing points in a two-dimensional diagram.
The numbers are used to label the crossings, and a superscript $+$ or $-$ to indicate an over-crossing or an under-crossing, respectively. 
For example, the Gauss code of the knot in Fig~\ref{fig:examples_knot}.a is 
$ 1^- \, 2^+ \, 3^- \, 1^+ \, 2^- \, 3^+$.
One of the most important properties for the knot representations is the possibility to map one representation $\mathcal{X}(K)$ to another $\mathcal{Y}(K)$~\cite{Lenhard2009}.


\subsection{Grid representation}


\begin{figure}
\centering

\begin{subfigure}{.3\linewidth}
  \centering
  \def\svgwidth{1\linewidth}
  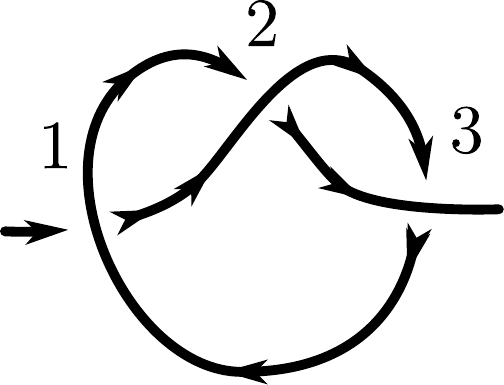
  \caption{}
  \label{fig:grid_overhand}
\end{subfigure}
\hfill
\begin{subfigure}{.3\linewidth}
  \centering
  \def\svgwidth{0.8\linewidth}
  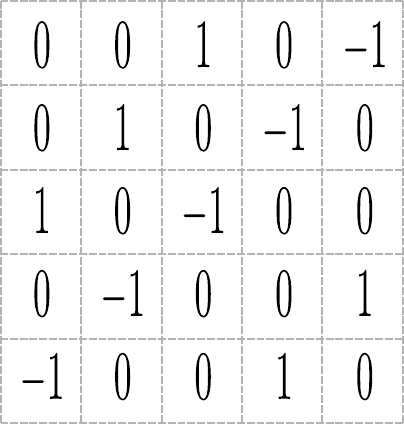
  \caption{}
  \label{fig:grid_grid}
\end{subfigure}
\hfill
\begin{subfigure}{.3\linewidth}
  \centering
  \def\svgwidth{\linewidth}
  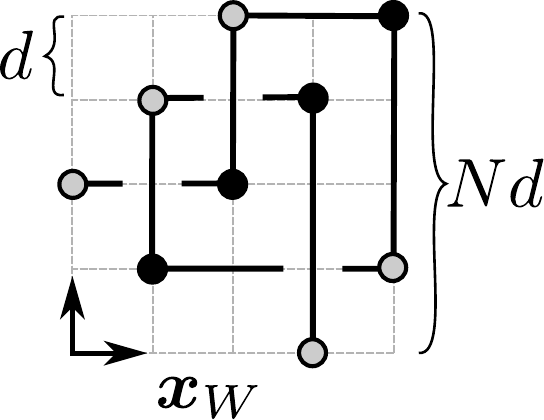
  \caption{}
  \label{fig:grid_points}
\end{subfigure}
\hfill

\caption{Knot representations: (a) a overhand knot, $c(K)=3$, b) Matrix of grid representation $N \times N$, c) Knot in 2D Cartesian coordinates.}
\label{fig:grid}
\end{figure}


Grid representation also called $N\times N$ grid diagram, arranges a knot based on vertical and horizontal lines between cells in the grid. 
The vertical lines cross the horizontal lines based on the number of crossings of the knot $c(K)$.
{Knots are homeomorphic, meaning that they can be deformed, but they will maintain the same topology. 
There is a {set} of knot representations that are homotopic among them, and there is a way to transform from one representation to another in the {set of equivalent representations}.
}
Mathematicians have shown that there is a transformation from any knot representation to a grid representation and vice versa \cite{Scherich2019ASO, Ozsvath2015GridHF}.
The content of each grid cell is either $1$, $-1$, or $0$. A grid diagram has the property that every row and column has exactly one $1$ and one $-1$. 
Then, we assemble the values to form a matrix associated with the knot, denoted by $\boldsymbol{A}=[a_{kj}]$, for $i,j\in\{1,...,N\}$.
Fig.~\ref{fig:grid_grid} shows the matrix~$\boldsymbol{A}$ of the overhand knot in Fig.~\ref{fig:grid_overhand}.
We open the knot to be able to fold it using a cable, so we can remove any number that does not have any crossing in its adjacent numbers. In the example of Fig.~\ref{fig:grid_grid}, we remove the left-bottom $-1$ and replace it with a $0$. We can always re-arrange the matrix by interchanging consecutive rows or columns while preserving the knot topology \cite{Scherich2019ASO}. 
So, we can assume that the knot always starts in the left bottom placement.
Then we can label the corners, i.e., $a_{kj}\neq 0$, in a sequence following the cable.
Each corner is transformed into a point $\boldsymbol{p}_i$, $i=\{1,...,n\}$.
Since each column has two points and we removed one point, the total number of points is $n = 2N - 1$.
Then, we can convert the corners to Cartesian coordinates. Let $\boldsymbol{p}_i$ be a point associated with a matrix cell $a_{kj}$, then, its coordinates are,
$$
\boldsymbol{p}_i= [d\cdot k,\: d\cdot j, z_p]^\top,
$$
where $z_p$ is the desired height where the plane of the grid is projected in $\mathbb{R}^3$,  $d$ is the desired cell width. The resultant points of the overhand knot are illustrated in Fig.~\ref{fig:grid_points}.




\subsection{The multi-catenary representation}


We find a curve composed of multiple connected catenary curves that are homotopic to the knot based on the grid representation.
The grid representation is a two-dimensional projection of the three-dimensional curve that describes the knot. Using the two-dimensional representation, we include the three-dimensional for each inflection of the knot held by the node.
We transform the knot into a multi-catenary curve based on the grid representation by converting every straight line into a catenary curve. We project the grid on the $xy$-plane at a coordinate $z_p$.
We enumerate every corner of the polyline and sequentially place the robots at each corner following the cable.
Since one cable passes over the other in the crossings, we want to maintain the topology and avoid self-intersections. For this purpose, we define two types of catenaries based on their height, {low and high} catenaries with height $h_{min}$ and $h_{max}$, respectively. The height of the horizontal plane where the points lie should satisfy $z_p>h_{max}$; otherwise, the cables would touch the floor without forming a catenary curve.



\begin{figure}[b]
    \centering
    \begin{subfigure}{.45\linewidth}
    \centering
    \includegraphics[trim=3.cm 1cm 3cm 1cm,clip,width=\linewidth]{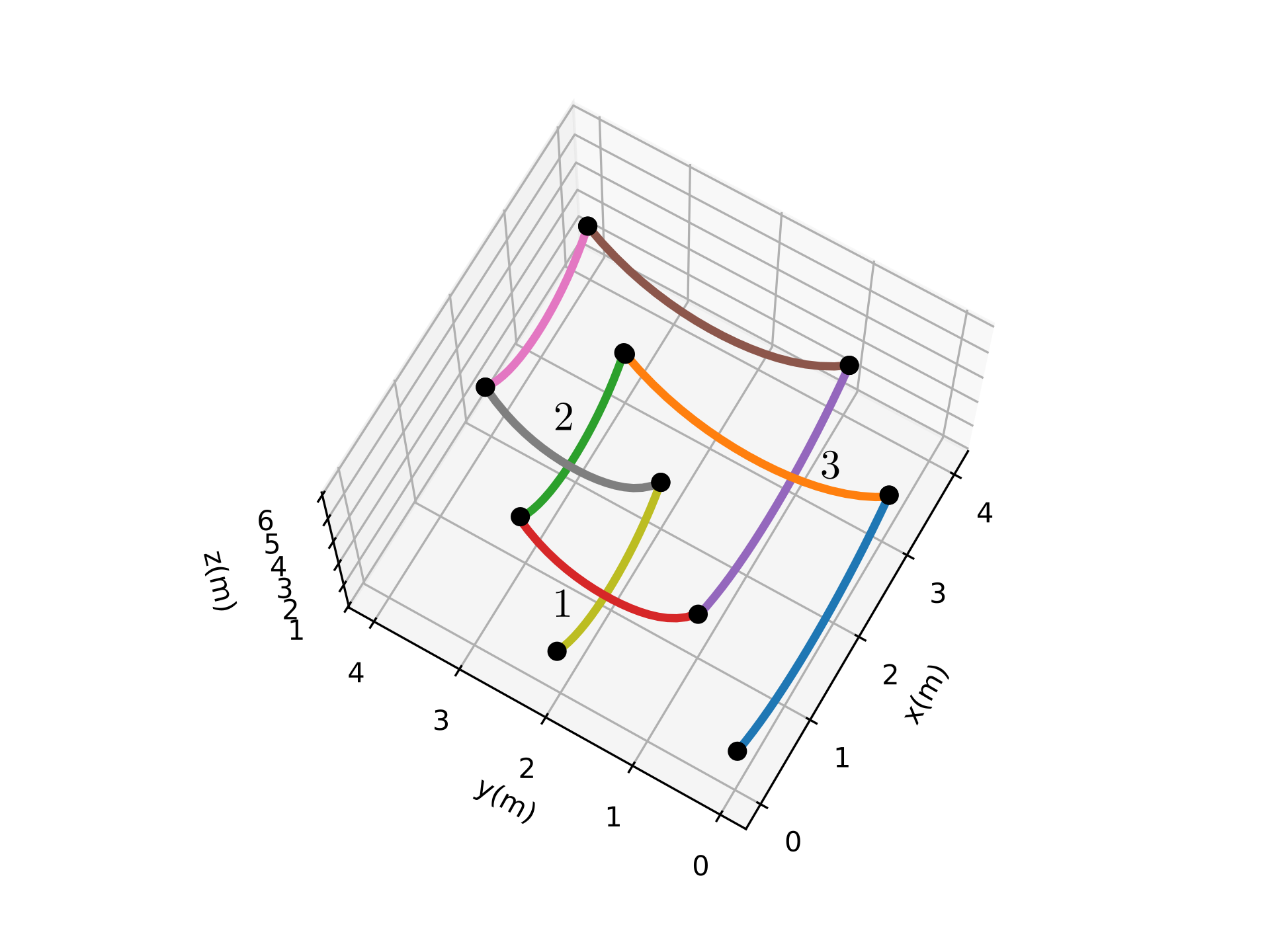}
    \caption{Overhand knot.}
    \end{subfigure}
    \hspace{.2cm}
    \begin{subfigure}{.45\linewidth}
    \center
    \includegraphics[width=\linewidth]{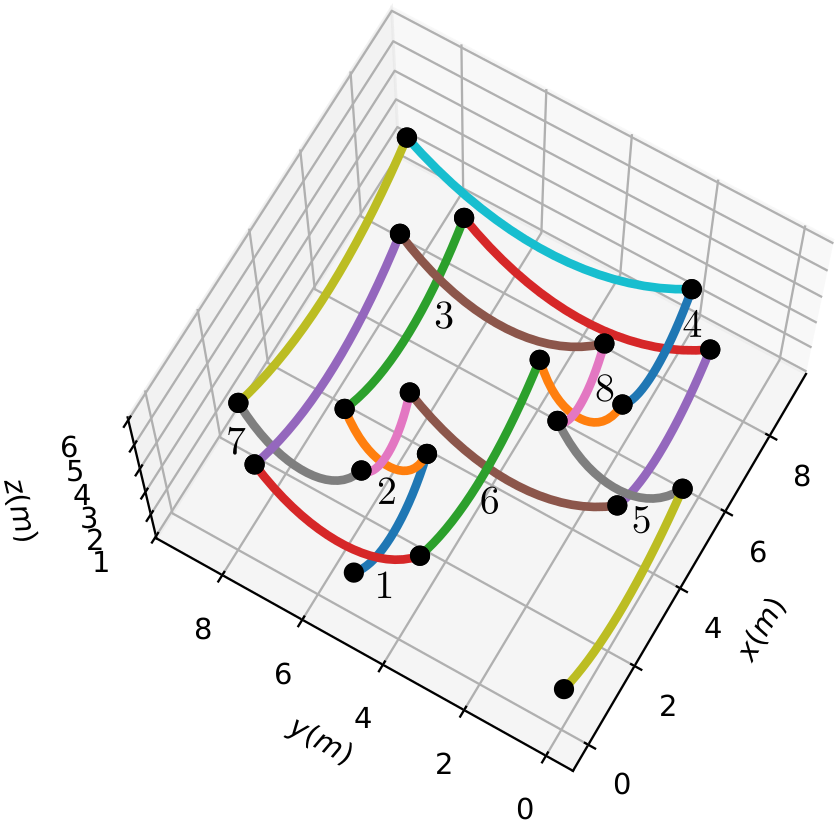}
    \caption{Carrick mat knot.}
    \end{subfigure}
    \caption{Knots constructed based on multiple catenary curves.}
    \label{fig:3D_knot_with_catenaries}
\end{figure}

The catenary curve has been widely studied, and its equations are well defined. 
The cable hanging from  two points forms a catenary curve \cite{lockwood1967book, routh_2013}, e.g. starting at point~$\boldsymbol{p}_i$ and ending at point~$\boldsymbol{p}_{i+1}$.
The general equation of a catenary curve in the $yz$-plane with height $h$, span $2s$, and lowest point at the origin is given by
\begin{equation}
\boldsymbol{\alpha}(r) = 
        \left\lbrack 
        \begin{array}{c}
            0\\
            r \\
            a \left (\cosh \frac{{r}}{a} - 1 \right)
        \end{array}
        \right\rbrack,
        \label{eq:catenarycurve}
\end{equation}
where $r\in [-s, s]$ is the curve parameter, and $a$ is a constant associated with the catenary shape. The constant $a\in \mathbb{R}_{\geq 0}$  can be obtained by numerically solving the following intrinsic equation,
\begin{equation}
     \frac{h}{a} =
    \cosh \left ( \frac{{s}}{a} \right ) -  1.
    \label{eq:sinh}
\end{equation}
%
Then, we can use the general equation to describe the catenary equation between the points $\boldsymbol{p}_i$ and $\boldsymbol{p}_{i+1}$.
The width, also called span, of the catenary is $2s_i=\|\boldsymbol{p}_{i+1}-\boldsymbol{p}_{i} \|$. 
There are two types of heights, $h_{min}$ and  $h_{max}$, for catenaries that pass above and below another catenary curve.
In the next section, we will find a value $h_{min}$ for a given $h_{max}$ such that the catenaries never self-intersect and maintain the knot topology.
We can numerically solve~\eqref{eq:sinh} to obtain the constant~$a_i$, and use it in~\eqref{eq:catenarycurve} to form
\begin{eqnarray}
\boldsymbol{\alpha}_{i}(r) &=& \boldsymbol{c}_{i} + \boldsymbol{R}_z(\theta_i) \boldsymbol{\alpha}(r), 
\end{eqnarray}
where $r\in [-s_i, s_i]$, and orientation of the catenary is $\theta_i~=~\pi (i-1)/2$, and $\boldsymbol{c}_{i}=(\boldsymbol{p}_{i}+\boldsymbol{p}_{i+1})/2 - [0,0,h_i]^\top$ is lowest point of the catenary curve.
The height of the catenary $h_i\in\{h_{min},h_{max}\}$ is  $h_{max}$ if it passes under another catenary in a crossing, and $h_{min}$ otherwise. Then we can describe the multi-catenary curve as a continuous piecewise function,
\begin{equation}
  \boldsymbol{\alpha}(t) =
    \begin{cases}
    \centering
      \boldsymbol{\alpha}_{1}(r), & \text{if  } 0 \leq r \leq q_2,\\
      \boldsymbol{\alpha}_{2}(r-q_2), & \text{if } q_2 < r \leq q_3,\\
      \quad \vdots &  \\
      \boldsymbol{\alpha}_{n-1}(r-q_{n-1}), & \text{if } q_{n-1} < r \leq q_{n},
    \end{cases}       
\end{equation}
where $q_k=\sum_j^{k-1} s_j$. We show examples of the multi-catenary representation in Fig.~\ref{fig:3D_knot_with_catenaries}.

\subsection{Total length and cable sections} 
Based on our multi-catenary curve, we can compute the length of each segment of the cable $\ell_i$.
The grid representation allows us to identify how to reconfigure the cable to reach our desired knot topology. 
For each cable segment between the points $\boldsymbol{p}_i$ and $\boldsymbol{p}_{i+1}$, we use the height $h_i$ and parameter $a_i$ in the equation of the catenary to obtain the length of the cable segment as $\ell_i=\sqrt{s_i^2 -a_i^2}$. Then, the total length of the cable is $\ell_T= \sum_{i=1}^{n-1}\ell_i$.

\subsection{Maintaining knot topology}
We need to guarantee that the multi-catenary curve does not self-intersect and maintains the knot topology $K$.
For a given $h_{min}$, we need to find an $h_{max}$ that guarantees that a cable segment that should be above another segment will keep like that independently of its span and location. 
To maintain the topology of the knot~$K$, every pair of curves $(i,j)$ in a crossing of the grid representation should satisfy the following condition. 

\textbf{Crossing condition:} In a grid representation, if a straight line between $\boldsymbol{p}_i$ and $\boldsymbol{p}_{i+1}$ passes over a straight line between $\boldsymbol{p}_j$ and $\boldsymbol{p}_{j+1}$, then in the multi-catenary representation, the catenary curve between $\boldsymbol{p}_i$ and $\boldsymbol{p}_{i+1}$ with height $h_{min}$ should always be above the catenary curve between $\boldsymbol{p}_j$ and $\boldsymbol{p}_{j+1}$ with height $h_{max}$.

The following proposition presents a lower bound for~$h_{max}$.


\begin{figure}
\centering
  \def\svgwidth{1\linewidth}
  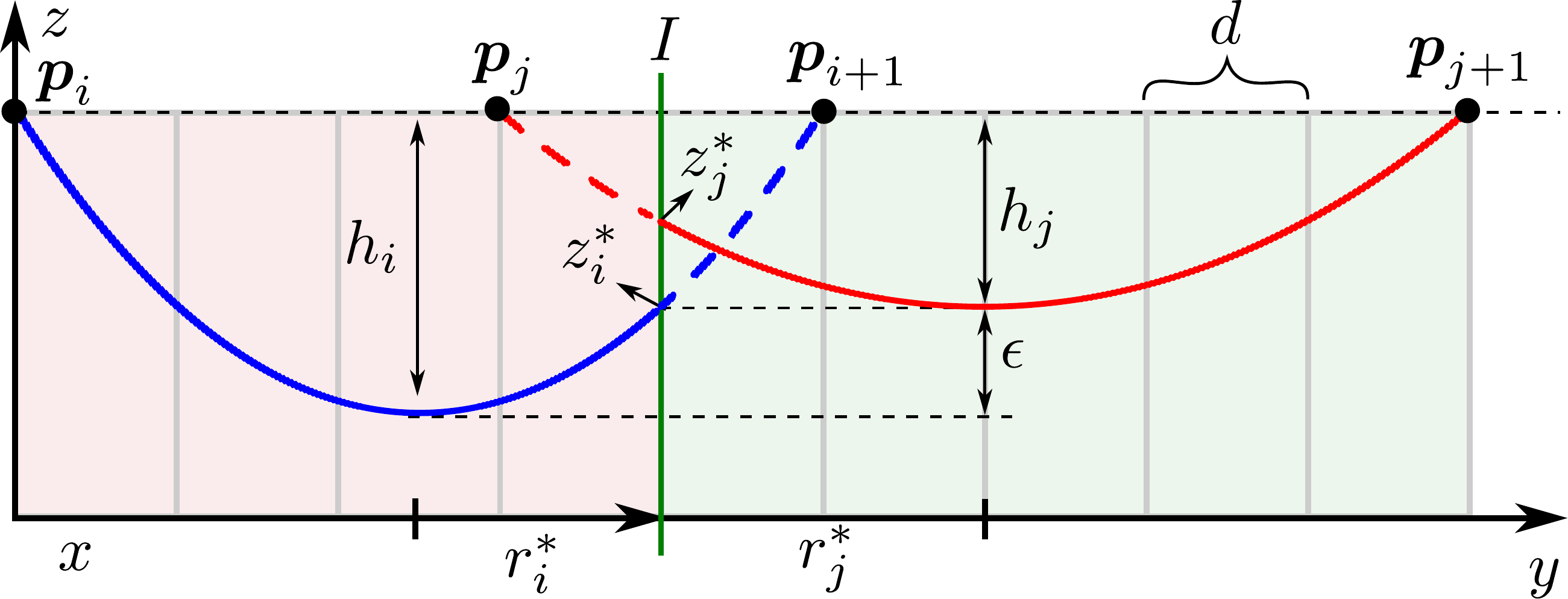
\caption{Crossing of two catenary curves in perpendicular planes. The red rectangle represents the $xz$-plane and the green rectangle represents the $yz$-plane. Both planes intersect at the vertical green straight line.}
  \label{fig:proof_intersection}
\end{figure}

\begin{proposition}
Given a multi-catenary representation, defined by a knot $K$ in a $N\times N$ grid representation with cell width $d$, and minimum catenary height $h_{min}$. 
The multi-catenary curve maintains the topology of $K$,  if $h_{max}$ satisfies the lower bound
\begin{equation}
h_{max}  > h_{min}  + \bar{a} \cosh\left( \frac{L-3d}{2\bar{a}} \right)+\bar{a},
\label{eq:hmin}
\end{equation}
where $\bar{a}$ is the constant of a catenary curve with height $h_{max}$ and span $s=(L-d)/2$.
\end{proposition}
\begin{proof}
To maintain the topology of the knot $K$, every pair of curves $(i,j)$ in a crossing of the grid representation has to satisfy the \emph{crossing condition}. 
Without loss of generality, we assume that the curve $i$ has height $h_{max}$ and lies on the $yz$-plane, and the curve $j$ has height $h_{min}$ and lies on the $xz$-plane. The other cases are covered by relabeling or swapping the planes.

Let $I$ be the vertical line where the planes of the curves intersect.
Since all curves start and end at the same height~$z_p$, we can use the equation of the catenary \eqref{eq:catenarycurve} to compute the~$z$ coordinates of the points where the curves intersect $I$,
\begin{eqnarray}
z_i^* &=& a_i \cosh{\left( \frac{r_i^*}{a_i}\right)} -a_i +z_p - h_i,\\
z_j^* &=& a_j \cosh{\left( \frac{r_j^*}{a_j}\right)} -a_j +z_p - h_j,
\end{eqnarray}
where $r_i^*$ and $r_j^*$ are the parameters of the curves $i$ and $j$ at the intersection with $I$ (see Fig. \ref{fig:proof_intersection}).
To hold the crossing condition, we must guarantee that the difference $z_j^*-z_i^*$ is always greater than zero. Let 
\begin{equation}
\epsilon=h_{max}-h_{min},
\label{eq:epsilon}
\end{equation}
and substitute $h_i=h_{max}$, $h_j=h_{min}=h_{max}-\epsilon$ in the difference, then  
\begin{equation}
z_j^*-z_i^* = a_j \cosh{\left(\frac{r_j^*}{a_j}\right) - a_j - a_i \cosh{\left(\frac{r_i^*}{a_i}\right) +a_i}} + \epsilon. 
\label{eq:deltaz}
\end{equation}
The worst-case scenario occurs when the curve $j$ reaches its minimum at $z_j^*$ by placing the minimum location of the curve at $I$, i.e., $r_j^*=0$; 
the curve $i$ reaches its maximum at $z_i^*$ when the span is maximized within the grid limits, $s=(L-d)/2$, and the distance to $I$ is minimized, in the grid case, one cell away, i.e., $r_i^*=s-d=(L-3d)/2$. We can use $s=(L-d)/2$ and $h_{max}$ in \eqref{eq:sinh} to obtain $a_i=\bar a$ in the worst case scenario. The value of $\epsilon$ in the worst case scenario is
\begin{equation}
\bar{\epsilon} = \bar a \cosh{\left(\frac{L-d}{2\bar a}\right) - \bar a}.
\label{eq:bound}
\end{equation}
We can verify that $z_j^*-z_i^*>0$ in the worst case scenario by choosing  any $\epsilon >\bar{\epsilon}$, and plugging
$r_j^*=0$, $r_i^*=s-d=(L-3d)/2$, $a_i=\bar a$ in \eqref{eq:deltaz}.
Finally, by replacing \eqref{eq:epsilon} and \eqref{eq:bound} in $\epsilon >\bar{\epsilon}$, we obtain the lower bound in \eqref{eq:hmin}.


\end{proof}





\section{Folding the Knot}
\label{section: Folding the knot}

Once we have a multi-catenary representation of the target knot $K$, we proceed to find a cable with the appropriate length, grab it at the appropriate locations, and proceed to fold it.
The three stages are the following.
\textit{i)} We use the total length of the curve in the multi-catenary representation~$\ell_T$ to define a sufficient length of the cable. This length is proportional to the cell width $d$ in the grid, so it is possible to re-scale the grid size if a cable with a fixed length is given.
\textit{ii)} Assuming that the cable starts completely extended, lying on the floor and forming a straight line. The robots grab the cable sequentially, respecting the distance $\ell_i$ between robots~$i$ and $i+1$.
\textit{iii)} To fold the knot, we use a leader-follower approach designing a single trajectory that all robots follow with a  proportional delay to their index. 
We summarize the folding process in Algorithm 1.

\begin{algorithm}
\textbf{Input:} Knot $K$, cell width $d$, Catenary height $h_{min}$, height of the vertical plane $z_p$.\\
\textbf{Output:} Trajectory function $\gamma(t)$

\caption{Fold a Knot}\label{alg:traj}
  \begin{algorithmic}[1]
    \State $N, \boldsymbol{A} \gets$ GridRepresentation($K$)
    \State $n,  \boldsymbol{p}_1,...,\boldsymbol{p}_n  \gets $ GridToPolyline($\boldsymbol{A},d$)
    \State $h_{max}, \ell_T, \ell_1,...,\ell_{n-1} \gets$ Multi-CatenaryRep($h_{min}, \boldsymbol{p}_i$)
    \State $\boldsymbol{\gamma}(t)\gets$ TrajectoryGen($h_{max},\boldsymbol{p}_i$)
    \State GrabCable($\ell_T, \ell_i$)
    \State FollowTrajectory($\boldsymbol{\gamma}(t), t_d, i$)
  \end{algorithmic}
\end{algorithm}

We denote the desired final locations by $\boldsymbol{p}_i^*$. Sending every robot $\boldsymbol{p}_i$ directly to $\boldsymbol{p}_i^*$ would entangle the cable and probably create a different knot. So the way robots reach their target points should be coordinated, respecting the crossing condition and maintaining the knot topology. So we define a trajectory $\boldsymbol{\gamma}(t)$ that is a curve parametrized by time $t$ and passes over all the target points $\boldsymbol{p}_i^*$.
Due to the sequential order, a leader-follower framework is suitable since the first robot (leader) will follow the trajectory, and the consecutive robots will follow the leader. The leader is the robot labeled as $i=n$.
The first robot starts at $t=0$ and the others start at time $t=t_d (n-i)$, where $t_d$ is the delayed time to start. Additionally, every robot $i$ will stop when it reaches its target point $\boldsymbol{p}_i^*$ on the trajectory.
Since the robots have to pass under and above the cables, we need to add intermediate waypoints to the current sequence $\boldsymbol{p}_1^*,..., \boldsymbol{p}_n^*$. The intermediate point between any pair of points $\boldsymbol{p}_i^*$ and $\boldsymbol{p}_{i+1}^*$ is
\begin{equation*}
    {\boldsymbol{q}}_i^* = \frac{\boldsymbol{p}_i^* +\boldsymbol{p}_{i+1}^*}{2} - \frac{2-(-1)^i}{2} h_{max} \hat{\boldsymbol{z}},\text{ for }i=1,...,n-1,
\end{equation*}
where $\hat{\boldsymbol{z}}=[0,0,1]^\top$.
Then, the new sequence of waypoint for the trajectory is $\boldsymbol{p}_1^*, \boldsymbol{q}_1^*,\boldsymbol{p}_2^*,..., \boldsymbol{q}_{n-1}^*, \boldsymbol{p}_n^*$. 
Then, we can create a trajectory by connecting all waypoints sequentially using a a piece-wise function defined by $n-1$ fifth-degree polynomials, also called Quintic functions. The control input for the robot is a trajectory-tracking controller,
$$
\boldsymbol{u}_i(t)= k_p(\boldsymbol{\gamma}(t_i)-\boldsymbol{p}_i(t)) +
k_d(\boldsymbol{\dot\gamma}(t_i)-\dot{\boldsymbol{p}}_i)(t) + \ddot{\boldsymbol{\gamma}}(t_i),
$$
where $t_i= t-t_d(n-i)$ if $t-t_d(n-i)>0$, otherwise robot $i$ waits. $t_d$ is the delay between robot $i$~and~$i+1$.
The trajectory is continuous, smooth and appropriate to be followed by our second order robot. One can use this trajectory to drive actual quadrotors \cite{mellinger2011}.





    


\section{Simulations}
\label{section: simulations}
We perform multiple simulations to test our Multi-Catenary robot and the knot folding method. We implement a numerical simulator in Python, and a realistic simulator in Mujoco 2.10 \cite{Todorov2012}.
For each knot, we start with its Gauss code and then apply Algorithm~\ref{alg:traj}.

\subsubsection{{Experiment 1--} Overhand knot}

The Gauss code is $K= 1^- \, 2^+ \, 3^- \, 1^+ \, 2^- \, 3^+$.
Our algorithm has the input $d=1$ and $h_{min}=1.34$.
The total execution time is 18.45 seconds. The knot matrix~$\boldsymbol{A}$ has a dimension of $5\times5$. Thus the number of robots is $n=9$. For the Overhand knot, the total length is $\ell_T = 18.83$ and the cable segments are $\{2.59, 2.44, 1.73, 2.87, 2.59, 2.44, 1.73, 2.44\}$.
To simulate the cable in Mujoco, we create a rope composed of 400 capsule objects; the distance for each link is $0.2$. We simulate an omnidirectional vehicle that holds the cable. The final configuration is shown in Fig~\ref{fig:EXP1_mujoco}.

\subsubsection{{Experiment 2--} Figure-eight knot}

\begin{figure}
\centering

\begin{subfigure}{.32\textwidth}
  \centering
  \def\svgwidth{0.8\textwidth}
  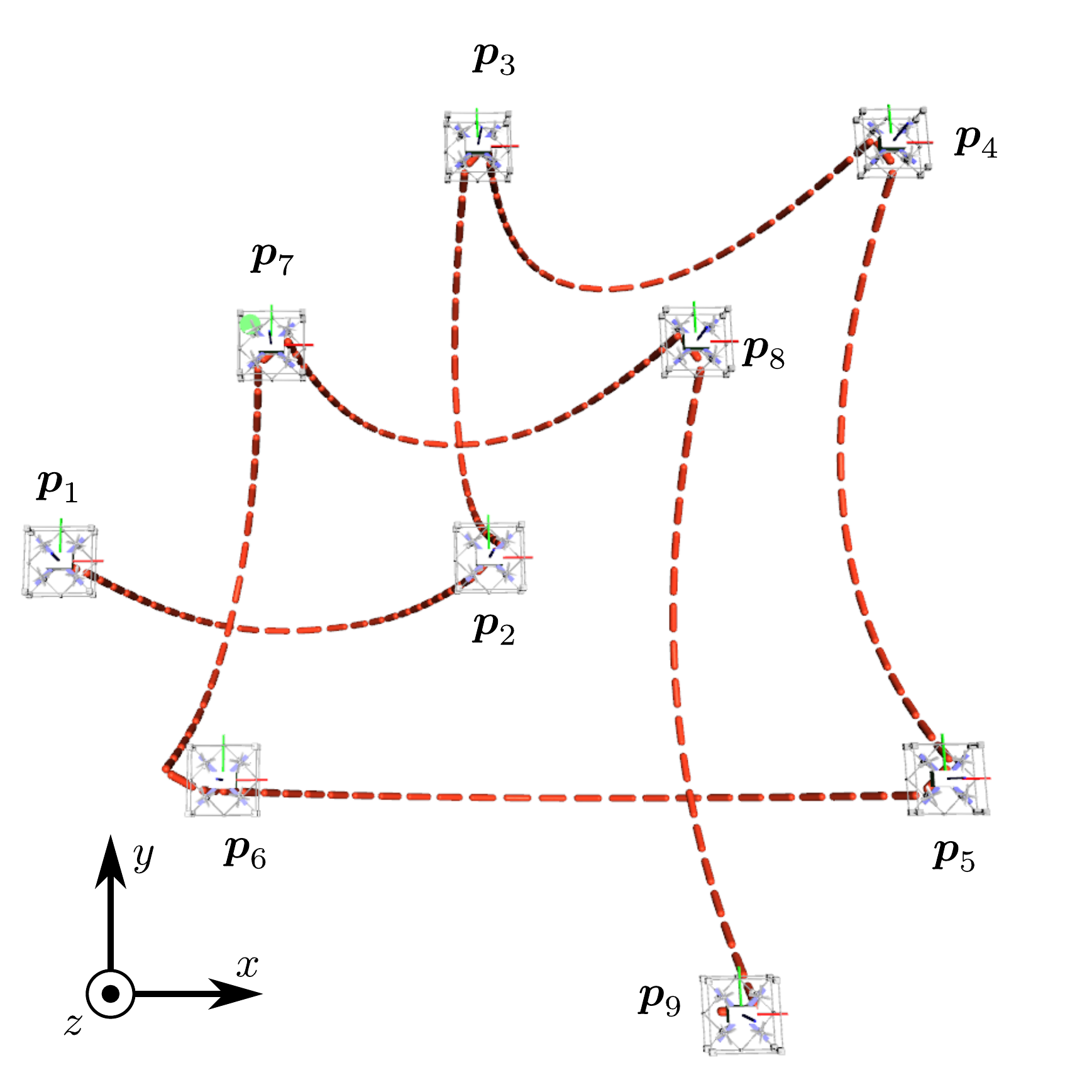
  \caption{Overhand knot.}
  \label{fig:EXP1_mujoco}
\end{subfigure}


\begin{subfigure}{.32\textwidth}
  \centering
  \def\svgwidth{0.8\textwidth}
  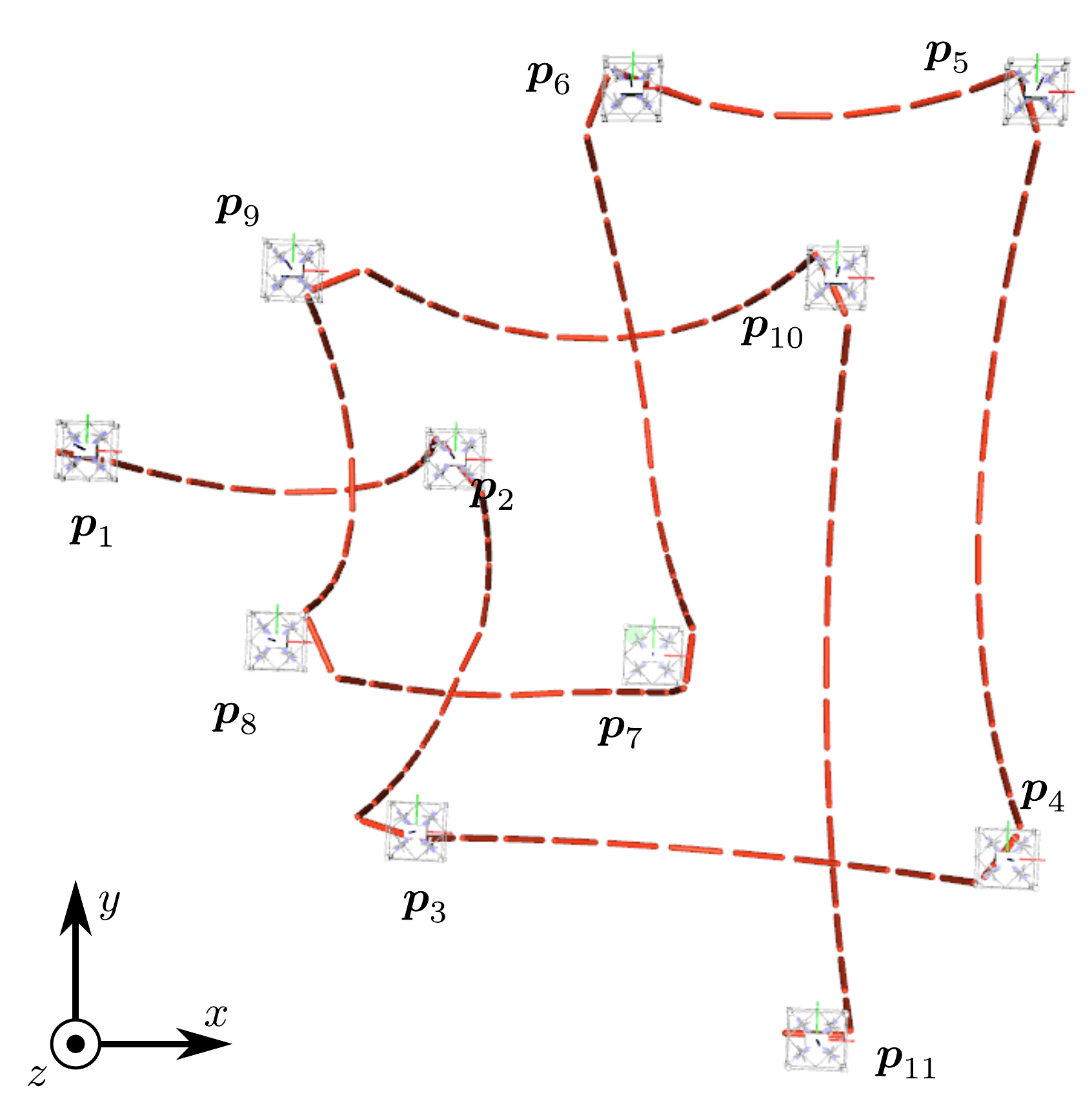
  \caption{Figure-eight knot.}
  \label{fig:EXP2_mujoco}
\end{subfigure}

\caption{Experiments 1 and 2: Knots constructed with catenaries in Mujoco.}
\label{fig:mujoco}
\end{figure}

We use a knot with Gauss code $K= 1^- \, 2^+ \, 3^- \, 4^+ \, 2^- \, 1^+ \, 4^- \, 3^+ $.
Our algorithm has the input $d=1$ and $h_{min}=1.55$.
The total execution time is 4680 seconds. The knot matrix~$\boldsymbol{A}$ has a dimension of $6\times6$, thus the number of robots are $n=11$. For the Figure-eight knot, the total length is $\ell_T = 52.48$ and the cable segments are $\{5.23, \, 5.98,\, 4.49,\, 4.74,\, 3.84,\, 5.32,\, 3.84,\, 5.98,\,$ $ 4.49, 4.74,\, 3.84\}$. To simulate the cable, we create a composite with 1101 capsule objects; the distance for each link is $0.2$. 
The final configuration is shown in Fig~\ref{fig:EXP2_mujoco}.

\subsubsection{{Experiment 3--}~Carrick mat knot}

\begin{figure}
\centering
\begin{subfigure}{.32\textwidth}
  \centering
  \includegraphics[trim=0.cm .5cm 0.cm 1.2cm,clip, width=\textwidth]{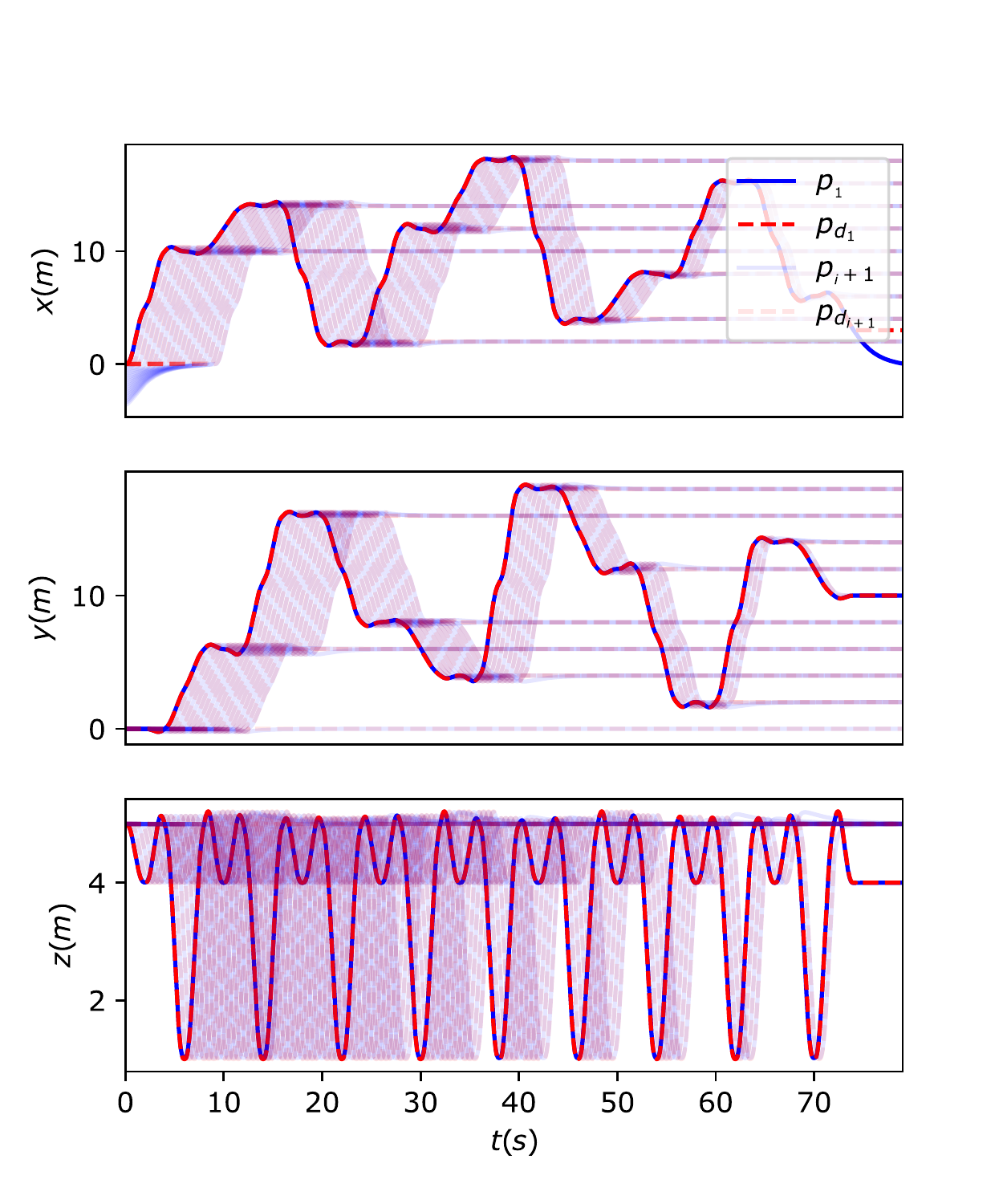}
  \caption{}
  \label{fig:EXP3_point_2D}
\end{subfigure}

\begin{subfigure}{.32\textwidth}
  \centering
  \includegraphics[trim=0.cm .3cm 0.cm 1.2cm,clip, width=\textwidth]{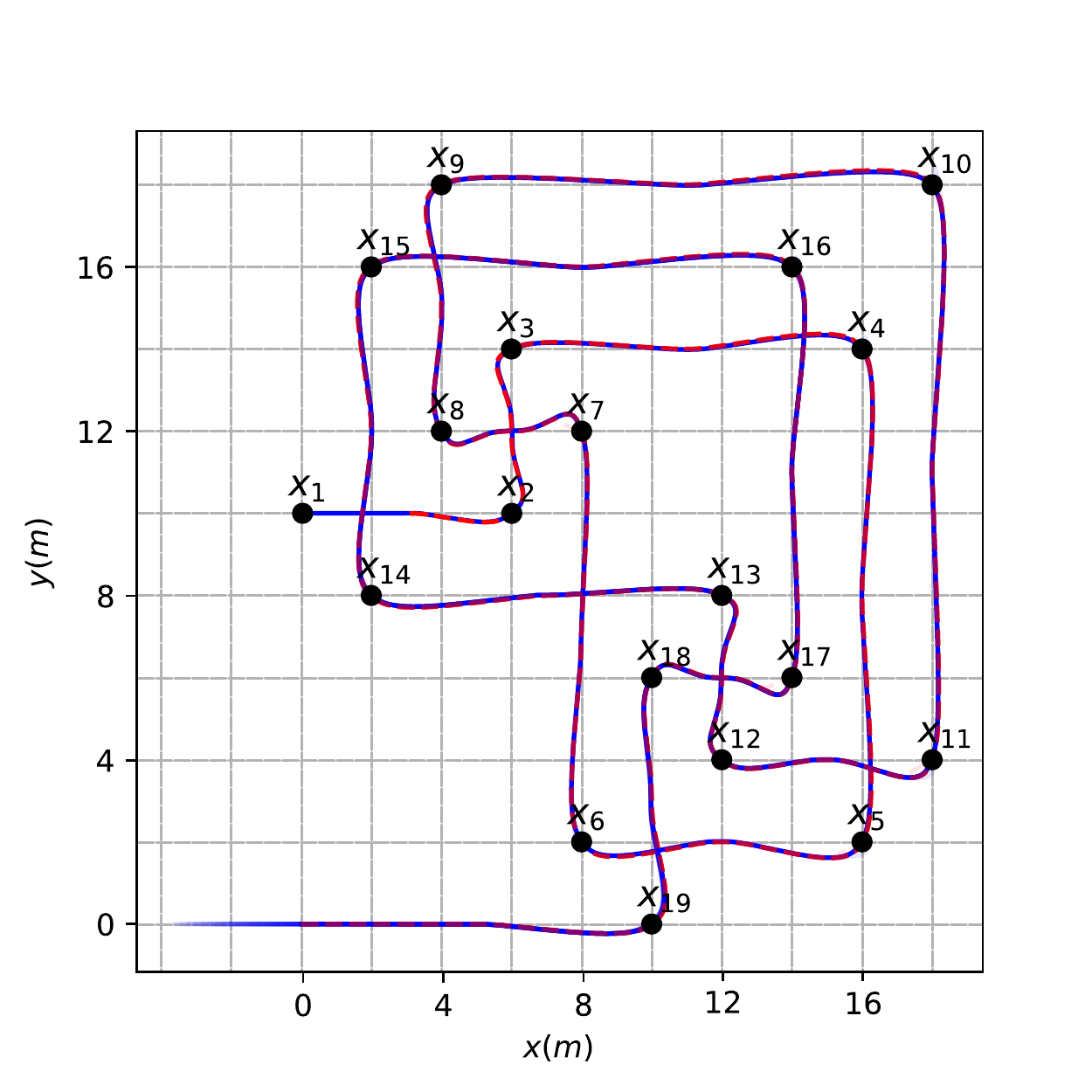}
  \caption{XY-plot.}
  \label{fig:EXP3_point_3D}
\end{subfigure}
\caption{Experiment 3: Carrick mat knot constructed with catenaries.}
\label{fig:pointrobot}
\end{figure}

The Gauss code is 
$K= 1^+ \, 2^- \, 3^+\, 4^- \, 5^+ \, 6^- \, 2^+ \, 7^- \, 4^+ \, 8^- \, 6^+ \, 1^- \, 7^+ \, 3^- \, 8^+ \, 5^-$.
The total length $\ell_T= 105.63$ and the cable segments are $\{4.74,\, 6.05,\, 7.51,\, 5.23,\, 6.72,\, 3.84,\, 5.32,\, 7.79,\, 8.34,\, 4.49,\, $ $ 4.74,\, 6.05,\, 5.98,\, 6.9,\, 6.72,\, $ $3.84,\, 5.32,\, 6.05\}$.
Our algorithm has the input $d=2$ and $h_{min}=2.89$.
Due to the large number of crossings of the knot, the simulator requires a longer cable with significantly more joints after discretization. 
Since the runtime of the solver in Mujocu increases exponentially to the number of joints, the simulation takes more than 10 hours to execute. Therefore, to test the scalability of our method, we develope a multi-catenary robot simulator in Python. We simulate the robots with second-order dynamics such that the catenaries can be computed in linear time. This numerical simulator, therefore, allows us to test our approach in real-time. 
The final configuration and the trajectories are shown in Fig~\ref{fig:pointrobot}.



\section{Conclusion and future work}
\label{section: conclusions}
We propose a method to fold knots in midair using a team of aerial robots. 
Based on a composition of catenary curves, we simplify the complexity of dealing with the infinite-dimensional configuration space of the cable, and formally propose a new knot representation.
This new representation allows us to design a trajectory that can be used to fold knots using a leader-follower approach. We show that our method works for different types of knots in simulations. 
In contrast to other simulators, our simulator simplifies the cable dynamics and runs in real-time for knots with many crossings.
Our solution is computationally efficient and can be executed in real-time.

In future work, we want to develop a faster method to fold the knots by parallelizing the robot actions. We also want to explore object transportation using knots and hitches.
The simplicity of cables and their lightweight make them a versatile tool with a great potential in aerial manipulation.





\bibliographystyle{ieeetr}
\bibliography{referencias.bib}

\end{document}